# Reliable Thermal Monitoring of Electric Machines through Machine Learning

Panagiotis Kakosimos
*ABB AB, Corporate Research*
Västerås, Sweden
panagiotis.kakosimos@se.abb.com

*Abstract*— The electrification of powertrains is rising as the objective for a more viable future is intensified. To ensure continuous and reliable operation without undesirable malfunctions, it is essential to monitor the internal temperatures of machines and keep them within safe operating limits. Conventional modeling methods can be complex and usually require expert knowledge. With the amount of data collected these days, it is possible to use information models to assess thermal behaviors. This paper investigates artificial intelligence techniques for monitoring the cooling efficiency of induction machines. Experimental data was collected under specific operating conditions, and three machine-learning models have been developed. The optimal configuration for each approach was determined through rigorous hyperparameter searches, and the models were evaluated using a variety of metrics. The three solutions performed well in monitoring the condition of the machine even under transient operation, highlighting the potential of data-driven methods in improving the thermal management.

I. INTRODUCTION

Electric machine failures happen for several reasons. Studies have shown that more than half of failures in induction machines are due to malfunctions in bearings, while most of the rest are caused by issues in the machine stator [1]. High temperatures can damage critical components such as stator/rotor windings and bearings. Therefore, more advanced cooling methods have been developed to prevent overheating from happening. A variety of methods available for monitoring the temperature at points inside electrical machines exist and can be classified into three main categories: sensor-based, model-based, and artificial intelligence (AI) algorithm-based [2], [3]. Each method has its advantages and disadvantages, but AI-based methods are becoming more appealing as they can quickly detect and diagnose problems in a generalized form. Additionally, these methods can also predict potential issues before they occur, allowing for preventive maintenance [4]. However, it is essential to highlight that, while AI-based methods can be very effective, they require a high level of expertise and resources to implement.

Machine learning (ML) has greatly enhanced the application of data-driven modeling on edge devices without requiring powerful processors, or cloud computing services and has minimal data transfer needs [5]. Several recent studies in the literature have used ML for monitoring the performance of electric powertrains including storage means, machines, and power converters [6]–[8]. Specifically, deep neural networks (NNs) have been used to predict the temperatures of permanent magnets in automotive applications and were found to be accurate under different operating scenarios [9]. Another approach proposed in the literature is using NNs to estimate the dynamic thermal behavior of a power chip and the rate of temperature change [10]. Additionally, a method to estimate the junction temperature of an IGBT power module using three non-linear ML-based models has also been proposed in [8], which has great prospects for industrial applications.

Moreover, in [7], ML was combined with traditional thermal modeling techniques such as Lumped Parameter Thermal Networks (LPTNs). This approach allowed the model to adapt its parameters and take environmental and aging effects into account, thereby enhancing its performance. Another study used ML to supervise battery health by considering multiple charging profiles during the training of the model [6], [11]. The use of temperature differences improved the performance of deep NNs in [12]. A neural network-based model was developed to estimate the internal temperatures of an induction motor in [13]. It is clear that AI/ML approaches offer many benefits and their applicability is being extended [14]. The knowledge of internal temperatures in a powertrain is crucial for various cases.

This paper presents an ML model for estimating internal machine temperatures. Several models were designed and trained using signals only available in electric drives enabling easy productization. The focus was on models that take continuous target variables as inputs, due to the nature of temperature signals. More specifically, a simplified approach, linear regression, and two more advanced ones, Convolutional Neural Networks (CNNs) and Recurrent Neural Networks (RNNs), were evaluated, and their performance in estimating internal temperatures was studied using an experimental test bench under varying conditions. The results showed that the developed ML models can estimate internal temperatures with high precision, opening up the possibility of implementing similar models for commercial applications. The models were able to detect an artificial failure injected to the machine's attached fan providing a solution for real-time temperature monitoring and enhanced reliability.

## II. SYSTEM SPECIFICATIONS

### A. Physical setup

A laboratory setup was utilized to produce datasets for training and evaluate the performance of the three investigated models in this study. Table I shows the specifications of the induction machine used in the experimental test bed. Various operating profiles were applied to the motor, each defining the speed and torque of the test subject under specific conditions. To create a diverse range of profiles, the dynamics and rate of change were varied and separated into three groups (i.e., slow, medium, and fast). This resulted in a total of eighteen profiles, providing roughly 150 hours of raw data for training and testing the developed models.

TABLE I. TEST MACHINE

| Motor | Test motor |
|---|---|
| Type | Induction machine |
| Power rating | 15 kW |
| Voltage rating | 400 V |
| Current rating | 30.6 A |
| Torque rating | 97 N m |
| Pole pair number | 2 |
| Speed | 1478 rpm |
| Cooling | Forced air |

During the tests, sensors were mounted on an induction motor to measure the temperature of the windings and drive-end (DE) and non-drive-end (NDE) bearings. This is a common practice in modern motors as they are equipped with such sensors. Temperature data were collected at a frequency sampling of 1 Hz. Additionally, other data such as current, voltage, speed, and others were also collected at the same rate. These parameters are important for providing the context of the machine's operation to the model. The data collected from these sensors were used to train and evaluate the performance of the two ML models.

### B. Preprocessing and feature expansion

Various signals were recorded during the machine operation in addition to the targeted temperatures of windings and bearings (PT100 sensors in the windings and outer ring of bearings). Analysis of the correlation matrix revealed that the motor speed and current were the most closely related to the thermal behavior of the motor. Additionally, the motor shell temperature, hereafter reference temperature, was found to be important as it provides the border conditions for the thermal model and the context of the environmental conditions. Table II provides a summary of all the inputs and outputs used in the model.

The next step in the process is to standardize the model inputs by transforming each feature into a normal distribution with a mean of zero and a unit variance. This step is essential as it makes the input contain both positive and negative values, which accelerates the learning process of the model. This step also ensures that the features are transformed into similar ranges, allowing the network to learn from each feature with equal effort improving the overall performance and robustness.

TABLE III. MEASURED INPUTS AND TARGET OUTPUTS

| Parameter | Symbol |
|---|---|
| Measured inputs | |
| Motor speed | $n_m$ |
| Motor current | $I_m$ |
| Reference temperature | $T_{ref}$ |
| Target temperatures | |
| Winding | $T_W$ |
| DE bearing | $T_{DE}$ |
| NDE bearing | $T_{NDE}$ |

Another pre-processing task that is performed is feature expansion with exponentially weighted moving averages for each selected feature at each time step. This technique provides linear models, which are not originally designed for handling sequential data, with sufficient references to the past machine behavior. On the other hand, the performance of CNNs and RNNs is also improved by minimizing the window length. Since the models need to have both short-term and long-term memories, eight different span values have been considered to achieve optimal performance of the models.

## III. MODEL OPTIMIZATION

For the model performance to be optimized, it is important to define the loss function and evaluation metrics.

### A. Loss function

A loss function, also known as a cost function, is used to evaluate how well a model fits the data. The function is calculated by comparing the model's estimations to the actual measurements [14]. There are many types of loss functions available in the literature, depending on the nature of the estimated targets. For example, loss functions used in classification problems differ from those used in regression problems. For regression problems, the most commonly used loss function is the mean squared error (MSE). Its well-known mathematical formula is as follows:

$$\text{MSE} = \frac{1}{N}\sum_{i=1}^{N}(Y_i - \hat{Y}_i)^2, \quad (1)$$

where $N$ is the number of samples, $Y$ is the measured value, and $\hat{Y}$ is the predicted one by the model. The objective of the loss function is to minimize the difference between the predicted and actual output. During the training process, the model's tunable parameters are optimized by minimizing the overall error. In this work, the method of gradient descent is used because of its effectiveness. Gradient descent works by moving in the direction of steepest decrease in the loss function, with the main formula as below:

$$a_{n+1} = a_n - \gamma \nabla L(a_n), \quad (2)$$

where $a$ is the tunable parameter to be updated, $L$ is the loss function to be minimized, and $\gamma$ impacts the step size of each update per each step. There are three main variants of gradient descent, which are batch, stochastic, and mini-batch gradient descents. One of the main differences between them is the amount of data each algorithm requires for each update. These variants can accelerate the optimization process, but they also affect the convergence rate. Therefore, choosing an appropriate schedule for adapting the learning rate is crucial for achieving optimal performance.

### B. Evaluation metrics

Once the hyperparameters of each model have been defined, the training process begins. During the process of training and testing, there are three different types of losses: training loss, validation loss, and testing loss. The training loss is calculated on the training set for each batch of data and is continuously updated during the training process. The validation loss, on the other hand, is calculated on a validation set for each epoch, which is separate from the training set. This loss is used to evaluate the model's performance during the training time and is used as a checkpoint for storing the optimal models and deciding when to stop training early. After the training process is completed, the model is evaluated on a final testing set, which is also separate from the training and validation sets. The testing loss is used to assess the final performance of the model.

There are various metrics used to evaluate the performance of the models. Mean squared error (MSE) is one of the most commonly used, as mentioned previously. Another metric is the Mean Absolute Error (MAE), which is similar to MSE with the only difference being that the square is replaced by the absolute value. MAE gives a more interpretable result than MSE. Although the error function is not differentiable in the whole space due to the absolute value, it can be used as a loss function with sub-gradient descent methods. On the other hand, the MSE loss function has the advantage of heavily penalizing large errors. Since large deviations in temperatures must be avoided, this loss function is a good candidate for this specific problem.

Another metric used to evaluate the performance of the models is the L-infinity norm, which finds the largest error for a set of estimations. This metric is significant in temperature estimation as it aims to keep the largest error small. It is important to note that some ML models may not have low MSE or MAE values, but if they have good L-infinity numbers, they can still be considered good candidates. Another metric that is often used in evaluating the performance of regression models is R-squared, also known as the coefficient of determination. It is calculated by comparing the sum of squares of residuals (the difference between the predicted values and the actual values) to the total sum of squares of the data. In general, it compares the estimated data set $(\hat{y_1}, \hat{y_2}, \hat{y_3}, \ldots, \hat{y_i})$ with the actual data set $(y_1, y_2, y_3, \ldots, y_i)$ with a mean of $\bar{y}$:

$$SS_{\text{res}} = \sum_i (y_i - \hat{y}_i)^2 = \sum_i e_i^2 \quad (3)$$

and the total sum of squares:

$$SS_{\text{tot}} = \sum_i (y_i - \bar{y})^2. \quad (4)$$

Then, R-squared is defined as follows:

$$R^2 = 1 - \frac{SS_{\text{res}}}{SS_{\text{tot}}}. \quad (5)$$

The coefficient of determination, or R-squared, is a metric that measures the percentage of variance that can be explained by the regression model in relation to the total variance. It is a normalized metric that ranges from zero to one, with a value of one indicating that the model perfectly explains the variation in the data and a value of zero indicating that the model does not explain any of the variations in the data. It is commonly used as a measure of goodness-of-fit for regression models.

## IV. Linear Regression

In this research, a linear model is considered as the baseline (Fig. 1). Despite its simplicity, it has the potential to be applied in real-world applications. A class called SGD Regressor from the scikit-learn library was used to test different linear models. The search space for the model includes the regularization multiplier parameter, mixing parameter as well as loss functions. Some loss functions when used with SGD Regressor can transform the linear model into a nonlinear one by specifying a kernel with epsilon insensitive loss function. However, this possibility was not explored in the hyperparameter search space in this research.

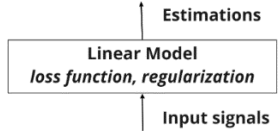

Fig. 1. Block scheme of a simplified linear model.

Linear regression considers the presence of a linear relationship between the scalar response and explanatory variables. If we consider $n$ observations $\{X_i, y_i\}_{i=1}^n$ where $X_i$ is a 1 by $p$ matrix, the relationship can be described as follows:

$$y_i = \beta_0 + \beta_1 x_{i1} + \beta_2 x_{i2} + \cdots + \beta_p x_{ip} + \epsilon_i, \quad (6)$$

where the term $\epsilon$ is used to model the unobserved random errors. In this approach, the term $\epsilon$ is used to model the unobserved random errors. The goal is to determine the weights and bias $\beta$ that minimize the loss function. The MSE is often used to compare the measured and predicted values. Due to the simplified nature of the method, the MSE loss function can be solved mathematically in a closed-form solution, without the need for gradient descent. This method is commonly known as the least square method, where $\beta$ is found as follows:

$$\hat{\beta} = (X^T X)^{-1} X^T Y. \quad (7)$$

The least square method provides an exact solution to the problem, however, it is not commonly used because it can be computationally intensive when dealing with large data sets. An alternative approach is to use the gradient descent method. In many applications, the trained model should be simple, meaning that it should not learn too many details from the data. This prevents overfitting and improves the generalizability of the model. In linear regression, regularization can help achieve this goal by adding penalty terms to the loss function and restricting the weights $\beta_i$ from becoming too large. In particular, elastic net regularization corresponds to the following:

$$\beta = argmin_\beta \left( ||y - X\beta||^2 + \lambda_2 ||\beta||^2 + \lambda_1 ||\beta||_1 \right), \quad (8)$$

where the first term is the MSE, the second quadratic penalty term is used in ridge regression, and the final term is found in lasso regression.

The loss function is a crucial aspect of ML models and its choice can affect the model's performance. The most common loss function used is the MSE, which is differentiable as already mentioned. Other commonly used loss functions include the epsilon insensitive loss function and the Huber loss function. The epsilon insensitive loss function does not penalize errors within a margin of tolerance, and beyond this margin, the common mean absolute error is used to determine the errors. This is referred to as hinge loss, and when it is combined with linear regression, the resulting model is the support vector regression. The choice of the loss function can have a significant impact on the model's performance and results.

$$L = \begin{cases} 0, & |y - f(X, \beta)| \leq \epsilon \\ |y - f(X, \beta)| - \epsilon, & otherwise. \end{cases} \quad (9)$$

The equation above shows the epsilon insensitive loss, where $\epsilon$ is a tunable parameter. However, the real power of support vector regression comes from the kernel method, which can handle nonlinear behaviors. The algorithm projects the data into higher dimensional spaces and then uses hyperplanes to approximate them. Another alternative loss function that is commonly used in robust regression is the Huber loss, it is less sensitive to outliers in the data set, and it performs better than the mean squared error loss function when the data set contains a lot of outliers. The Huber loss function has the following form:

$$L = \begin{cases} \frac{1}{2}(y - f(X, \beta))^2, & |y - f(X, \beta)| \leq \delta \\ \delta\left(y - f(X, \beta) - \frac{1}{2}\delta\right), & otherwise. \end{cases} \quad (10)$$

## V. Convolutional NNs

The CNN model is different from the linear model in that it requires input data to be in a sequential form. The model predicts the output at a future time step by using a sequence of input data. The sequence length is a hyperparameter that must be fine-tuned. To reduce the amount of data needed for one prediction, the model uses a Global Pooling layer to downsample the inputs horizontally, and then a dense layer performs the regression.

This method of using a sequence of input data and fine-tuning the sequence length is specific to CNN models.

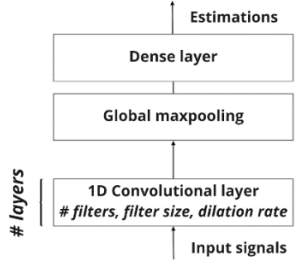

Fig. 2. A typical block scheme of convolutional NNs.

In the context of image processing, CNN layers are often used as feature extraction layers, while the dense layer performs the regression or classification based on the extracted features. CNN layers can also be stacked, and the number of layers and filters, filter sizes on each layer, and dropout and dilation rates can be defined based on the application's needs (Fig. 2).

In this work, the targets are sequential temperatures, making neural networks that receive sequential data a better fit than simplified fully connected neural networks. Convolutional neural networks are thus good candidates. While 2D CNNs are used in image processing applications due to their capability to explore spatial data, the 1D variant is a good candidate for exploring temporal data. Fig. 3 shows the basic structure of such a network. The first layer is the input layer, whereas the last layer is the output layer. Several hidden layers may exist and contain different numbers of neurons and connections from previous layers. An activation function is assigned to each neuron to attach a non-linear behavior to the network.

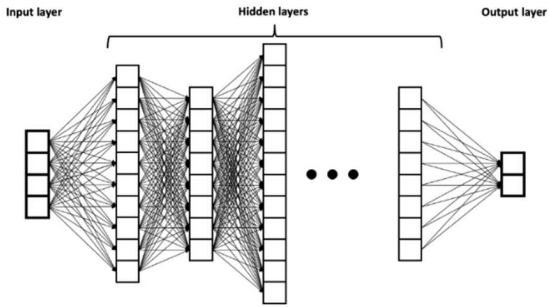

Fig. 3. Basic structure of a CNN model.

In a temporal CNN, the data comes in a sequential form, meaning that one neuron at the input layer represents the entire input data at one time step, regardless of its dimension. Adjacent neurons represent inputs from adjacent time steps. Each neuron not only has information from the current time step but also has information from previous inputs. If more convolutional layers are stacked, neurons at the latter layers get information from inputs that originate from a long time in the past. This allows CNNs to incorporate temporal information in their predictions.

## VI. RECURRENT NNs

Another network structure particularly capable of dealing with sequential data is the recurrent neural network (RNN). The RNN model has a similar form to CNNs with multiple RNN layers followed by a global pooling layer and a dense layer regressor (Fig. 4). The main principle of RNN is that the prediction made at each time step not only depends on the current inputs but also on the prediction of previous steps. Nevertheless, by solely altering CNNs and adapting them into RNNs would not work due to exploding or vanishing gradient problems. Two good candidates that tackle these known problems are long short-term memory (LSTM) and gated recurrent units. The gated recurrent unit combines several gates and states into one thus needing less parameters with faster training speeds. However, LSTM outperforms most of such variants in terms of accuracy, so in this work, RNN refers to the long short-term memory block.

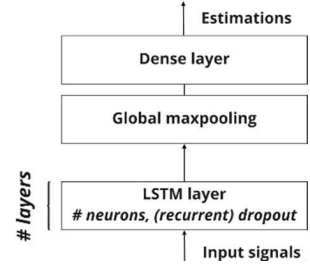

Fig. 4. A typical block scheme of recurrent NNs.

In particular, if $h$ is the output and $x$ is the input of a typical LSTM cell, the cell takes the output from the previous time step, $h_{t-1}$, together with the current input vector $x_t$ as inputs. In addition to those two parameters and in order to track long-term memory, the cell also maintains a state vector, $c$. The cell uses or updates $c$ at each time step. Inside the cell, there are three gates: an input, a forget, and an output gate. Each gate performs tasks as in other typical neural schemes, that is to take the linear combination of the inputs and apply an activation function:

$$\begin{aligned} f_t &= \sigma_g(W_f x_t + U_f h_{t-1} + b_f) \\ i_t &= \sigma_g(W_i x_t + U_i h_{t-1} + b_i) \\ o_t &= \sigma_g(W_o x_t + U_o h_{t-1} + b_o) \\ \tilde{c}_t &= \sigma_c(W_c x_t + U_c h_{t-1} + b_c) \\ c_t &= f_t \circ c_{t-1} + i_t \circ \tilde{c}_t \\ h_t &= o_t \circ \sigma_h(c_t) \end{aligned} \quad (11)$$

Afterward, multiple LSTM cells can be stacked, where $x$ is fed into the cells sequentially, and $h$ is generated in sequence, too. By stacking multiple LSTM layers, $h_{t-1}, h_t, h_{t+1}, \ldots$ are treated as inputs to the other LSTM layers. A disadvantage of the LSTM scheme is that the training process cannot be executed in parallel. For example, the second LSTM cell cannot calculate its parameters without the output of the first cell, makingn makes the training time of an LSTM layer particularly long. Furthermore, due to its inherent sophisticated structure of each cell, the same length of an LSTM layer consists of more parameters than CNN layers. Finally, the hyperparameters that need to be tuned are the number of layers and neurons in each layer, the dropout rates, and recurrent dropout rates.

VII. RESULTS AND DISCUSSION

In this section, the three ML models that were investigated, along with their best configurations determined through hyperparameter search, have been used to predict the machine temperatures under a representative operating profile. This allows for a comparison of their performance in estimating the internal temperatures of the motor under the same conditions. By using the best configurations, it can be determined which model performs the best in this specific scenario and how it compares to the other model.

*A. Hyperparameter search*

With all the necessary information available, the models are ready to be trained. However, before finalizing the parameters of each model, it is essential to run a hyperparameter optimization. This process involves fine-tuning a lot of parameters and find the best configuration for the specific task each time. The final selected model configurations are summarized in Table II, and they will be used to train the final models and make predictions. This table provides a summary of the best set of parameters for each model, which have been obtained through the hyperparameter search process.

TABLE II. MODEL OPTIMAL PARAMETERS

| Symbol | Hyperparameter | Value |
|---|---|---|
| | Linear model | |
| $\alpha$ | Penalty coefficient | 0.43 |
| $l1$ | Mixing parameter | 0.99 |
| $loss$ | Loss function | Squared error |
| | CNN (3 layers) | |
| $n_{filter}$ | No. filters | 125,5,125 |
| $s_{filter}$ | Filter sizes | 2,2,2 |
| $d$ | Dilation rates | 3,1,1 |
| $l$ | Sequence length | 100 |
| | RNN (2 layers) | |
| $n$ | No. Neurons | 22,72 |
| $\beta_d$ | Dropout rates | 0,0.1 |
| $\beta_r$ | Recurrent Dropout rates | 0.4,0.1 |
| $l$ | Sequence length | 100 |

*B. Operating profiles*

To evaluate the model performance thoroughly, all generated profiles have been used as test data, at least once. This was done by removing each profile from the training set one by one and training the model on the remaining profiles. Additionally, during the training process, another small set of profiles was set aside as the validation set. This approach allows for the model to be tested on a diverse set of data and ensures that the model has a good generalization performance. This technique of using different profiles as test data at least once and using the validation set during training is commonly used to evaluate model performance more thoroughly and realistically. One profile that includes both slow and fast dynamics has been used for demonstrating the model performance (Fig. 5).

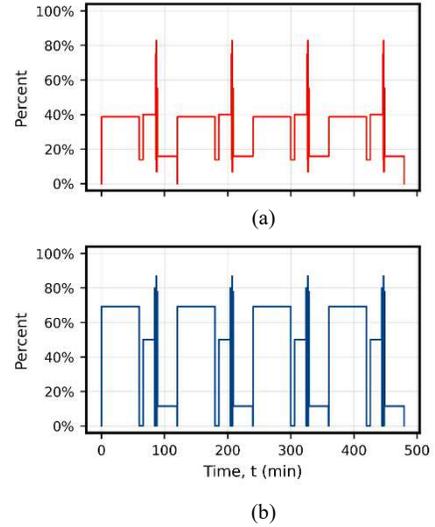

Fig. 5 Speed (a) and torque (b) of the test machine under one operating profile.

*C. Linear model*

A penalty coefficient of 0.43 and a mixing parameter of 0.99 were chosen by the hyperparameter search for the linear model. Due to the high mixing parameter, the model utilizes L1 regularization. Fig. 6 illustrates the performance of the model that was trained to predict the machine targeted temperatures. The top row displays the measured and predicted temperatures, while the second row presents the errors for each target. The bottom row serves as a reference, displaying the inputs for this specific profile. The linear model has an MSE of 1.94 for this operating profile. Overall, the model's performance is judged satisfactory. The majority of the time, the error is under 4ºC, which is suitable for most commercial applications. The model performs exceptionally well for the NDE bearing temperature, with minimal errors. There is no significant difference in performance during the first half of the profile (slow dynamics) and the second half (fast dynamics).

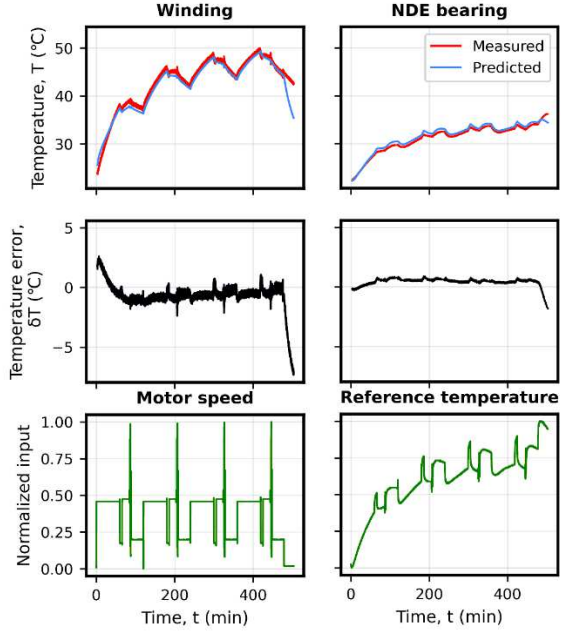

Fig. 6. Linear model performance.

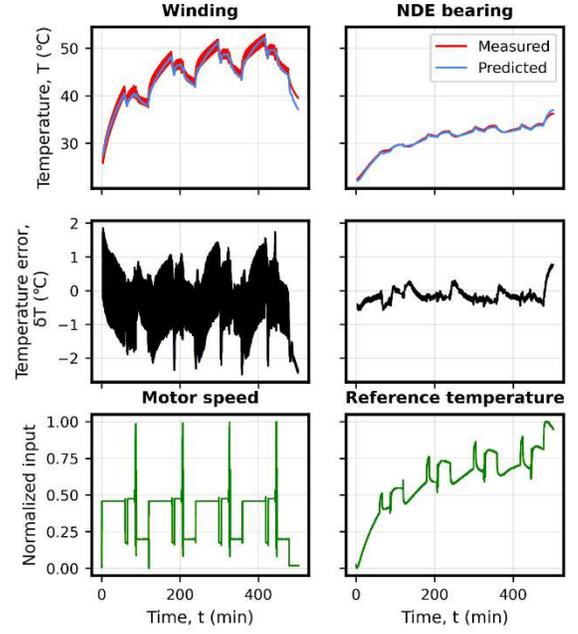

Fig. 7. CNN model performance.

*D. CNNs*

As a sequential model, the CNN model is expected to quickly learn from the sequential nature of the temperature data. However, it is also more complex and larger than the linear models. During the hyperparameter search, the optimal number of layers was three. The search revealed that relatively small filter sizes and a high dilation rate on the first layer were optimal. The results of the CNN model under the same operating profile of the linear model can be seen in Fig. 7. The MSE for this model is 0.54, and its predictions of motor targets were slightly less accurate than the linear model. Most of the time, the temperature error remained below 3°C. Due to the large thermal inertia of machines, linear models can also perform satisfactorily, however, under more dynamic conditions, CNNs are expected to outperform conventional models.

*E. RNNs*

The RNN model is also a sequential model able to tackle higher levels of nonlinearities due to its sophisticated structure. The hyperparameter search revealed that the optimal number of layers was two for this model. The results of the CNN model under the same operating profile of the linear model can be seen in Fig. 8. The MSE for this model is 0.51, and its predictions of motor targets were marginally more accurate than the linear and CNN models. The absolute temperature error remained at low levels during both slow and fast dynamics.

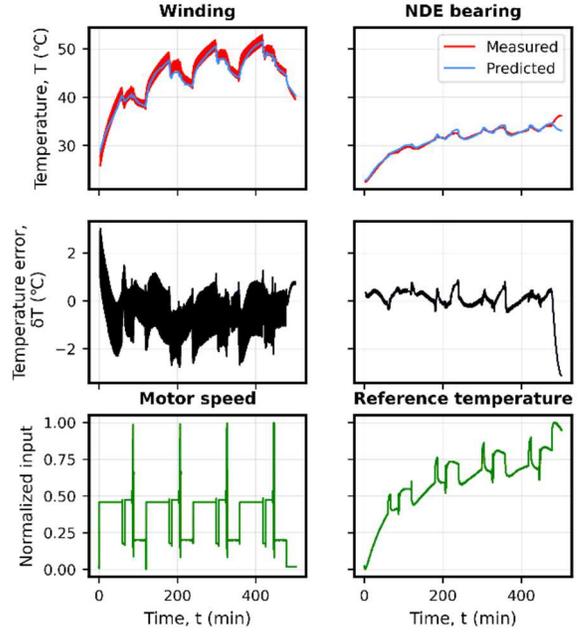

Fig. 8. RNN model performance.

*F. Condition monitoring of cooling performance*

The benefits that derive from estimating internal machine temperatures with high accuracy can be various. One of the most important aspects is the reliability enhancement of the operation of electrical machines. For this purpose, an artificial failure has been injected into the machine by blocking the fan surface by about 70% after 3 hours of operation, as shown in Fig. 9.

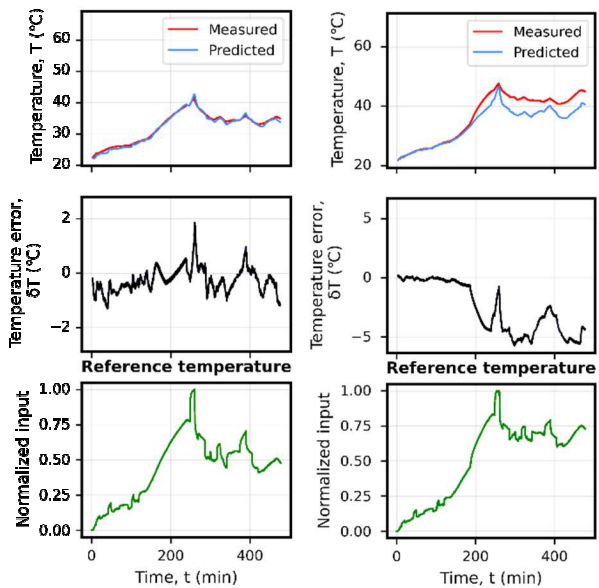

Fig. 9. Measured and predicted NDE bearing temperatures under healthy (left) and faulty (right) conditions with RNNs.

The deviation of the measured and predicted bearing temperatures reveals the impact of the blocked fan (Fig. 9). During the healthy operation, the RNN model tracks the temperature changes; however, at the unexpected event of the blocked fan, the two signals deviate significantly. Due to the large thermal inertia, the phenomenon propagates gradually, and the error exceeds that of the normal operation signaling an alert. The model was trained only under healthy conditions; thus, any deviations from the training dataset are highlighted.

## VIII. Conclusion

In this study, three ML approaches was applied to transform the way internal motor temperatures are estimated. Through rigorous testing of three ML algorithms, from simple linear regression to state-of-the-art deep neural networks, it was possible to achieve high accuracy in predicting the temperatures of the stator winding and bearings. Utilizing experimental lab data obtained by running the system under various operating conditions, including both slow and fast dynamics, a thorough hyperparameter search for each model was conducted to identify its optimal configuration. The results of the model evaluations, using multiple performance metrics, were promising: ML models emerged as an approach that can deliver exceptional performance even under high dynamic and rapidly changing conditions with the possibility of being easily generalized. Under harsh cooling conditions, the RNN model was able to detect deviations from a normal operation, thus allowing a more reliable detection of abnormal conditions.